\def\year{2018}\relax
\documentclass[letterpaper]{article}
\usepackage{aaai18}
\usepackage{times}
\usepackage{helvet}
\usepackage{courier}
\usepackage{amsmath}
\usepackage{amsfonts}
\usepackage{bbm}
\usepackage{enumitem}
\usepackage{graphicx}
\usepackage{multirow}
\usepackage{subcaption}
\usepackage{xcolor}
\definecolor{darker}{rgb}{0,0.15,0.8}
\usepackage{appendix}
\usepackage{url}
\usepackage[colorlinks, urlcolor=darker, citecolor=darker, linkcolor=darker]{hyperref}

\newcommand*\rot{\rotatebox{90}}

\DeclareMathOperator*{\argmax}{argmax}
\newcommand{\oneS}{\ensuremath{{}^{\textstyle *}}}
\newcommand{\twoS}{\ensuremath{{}^{\textstyle **}}}

\title{DeepType: Multilingual Entity Linking by Neural Type System Evolution}

\begin{document}
%
\author{Jonathan Raiman\\
OpenAI\\
San Francisco, California\\
\texttt{raiman@openai.com} \\
\And
Olivier Raiman\\
Agilience\\
Paris, France\\
\texttt{or@agilience.com}
}

\maketitle
\begin{abstract}
The wealth of structured (e.g. Wikidata) and unstructured data about the world available today presents an incredible opportunity for tomorrow's Artificial Intelligence. So far, integration of these two different modalities is a difficult process, involving many decisions concerning how best to represent the information so that it will be captured or useful, and hand-labeling large amounts of data.
DeepType overcomes this challenge by explicitly integrating symbolic information into the reasoning process of a neural network with a type system.
First we construct a type system, and second, we use it to constrain the outputs of a neural network to respect the symbolic structure. We achieve this by reformulating the design problem into a mixed integer problem: create a type system and subsequently train a neural network with it. In this reformulation discrete variables select which parent-child relations from an ontology are types within the type system, while continuous variables control a classifier fit to the type system. The original problem cannot be solved exactly, so we propose a 2-step algorithm: 1) heuristic search or stochastic optimization over discrete variables that define a type system informed by an Oracle and a Learnability heuristic, 2) gradient descent to fit classifier parameters.
We apply DeepType to the problem of Entity Linking on three standard datasets (i.e. WikiDisamb30, CoNLL (YAGO), TAC KBP 2010) and find that it outperforms all existing solutions by a wide margin, including approaches that rely on a human-designed type system or recent deep learning-based entity embeddings, while explicitly using symbolic information lets it integrate new entities without retraining.
\end{abstract}

\section{Introduction}

Online encyclopedias, knowledge bases, ontologies (e.g. Wikipedia, Wikidata, Wordnet), alongside image and video datasets with their associated label and category hierarchies (e.g. Imagenet~\cite{deng2009imagenet}, Youtube-8M~\cite{abu2016youtube}, Kinetics~\cite{kay2017kinetics}) offer an unprecedented opportunity for incorporating symbolic representations within distributed and neural representations in Artificial Intelligence systems.
Several approaches exist for integrating rich symbolic structures within the behavior of neural networks: a label hierarchy aware loss function that relies on the ultrametric tree distance between labels (e.g. it is worse to confuse sheepdogs and skyscrapers than it is to confuse sheepdogs and poodles) \cite{wu2017hierarchical}, a loss function that trades off specificity for accuracy by incorporating hypo/hypernymy relations \cite{deng2012hedging}, using NER types to constrain the behavior of an Entity Linking system \cite{ling2015design}, or more recently integrating explicit type constraints within a decoder's grammar for neural semantic parsing \cite{jayant2017parsing}.
However, current approaches face several difficulties:
\begin{itemize}
\item Selection of the right symbolic information based on the utility or information gain for a target task.
\item Design of the representation for symbolic information (hierarchy, grammar, constraints).
\item Hand-labelling large amounts of data.
\end{itemize}

DeepType overcomes these difficulties by explicitly integrating symbolic information into the reasoning process of a neural network with a type system that is automatically designed without human effort for a target task. We achieve this by reformulating the design problem into a \text{mixed integer} problem: create a type system by selecting roots and edges from an ontology that serve as types in a type system, and subsequently train a neural network with it. The original problem cannot be solved exactly, so we propose a 2-step algorithm:
\begin{enumerate}
\item heuristic search or stochastic optimization over the discrete variable assignments controlling type system design, using an Oracle and a  Learnability heuristic to ensure that design decisions will be easy to learn by a neural network, and will provide improvements on the target task,
\item gradient descent to fit classifier parameters to predict the behavior of the type system.
\end{enumerate}

In order to validate the benefits of our approach, we focus on applying DeepType to Entity Linking (EL), the task of resolving ambiguous mentions of entities to their referent entities in a knowledge base (KB) (e.g. Wikipedia).
Specifically we compare our results to state of the art systems on three standard datasets (WikiDisamb30, CoNLL (YAGO), TAC KBP 2010). 
We verify whether our approach can work in multiple languages, and whether optimization of the type system for a particular language generalizes to other languages\footnote{e.g. Do we overfit to a particular set of symbolic structures useful only in English, or can we discover a knowledge representation that works across languages?} by training our full system in a monolingual (English) and bilingual setup (English and French), and also evaluate our Oracle (performance upper bound) on German and Spanish test datasets. We compare stochastic optimization and heuristic search to solve our mixed integer problem by comparing the final performance of systems whose type systems came from different search methodologies. We also investigate whether symbolic information is captured by using DeepType as pretraining for Named Entity Recognition (NER) on two standard datasets (i.e. CoNLL 2003 \cite{Sang2003IntroductionTT}, OntoNotes 5.0 (CoNLL 2012) \cite{Pradhan2012CoNLL2012ST}).

Our key contributions in this work are as follows:
\begin{itemize}
\item  A system for integrating symbolic knowledge into the reasoning process of a neural network through a type system, to constrain the behavior to respect the desired symbolic structure, and automatically design the type system without human effort.
\item An approach to EL that uses type constraints, reduces disambiguation resolution complexity from $O(N^2)$ to $O(N)$, incorporates new entities into the system without retraining, and outperforms all existing solutions by a wide margin.
\end{itemize}
We release code for designing, evolving, and training neural type systems\footnote{\url{http://github.com/openai/deeptype}}. Moreover, we observe that disambiguation accuracy reaches 99.0\% on CoNLL (YAGO) and 98.6\% on TAC KBP 2010 when entity types are predicted by an Oracle, suggesting that EL would be almost solved if we can improve type prediction accuracy.

The rest of this paper is structured as follows. In Section~2 we introduce EL and EL with Types, in Section~3 we describe DeepType for EL, In Section~4 we provide experimental results for DeepType applied to EL and evidence of cross-lingual and cross-domain transfer of the representation learned by a DeepType system. In Section~5 we relate our work to existing approaches. Conclusions and directions for future work are given in Section~6.

\section{Task}

Before we define how DeepType can be used to constrain the outputs of a neural network using a type system, we will first define the goal task of Entity Linking.
\paragraph{Entity Linking} The goal is to recover the ground truth entities in a KB referred to in a document by locating mentions (text spans), and for each mention properly disambiguating the referent entity. Commonly, a lookup table that maps each mention to a proposal set of entities for each mention $m$: $\mathcal{E}_m=\{e_1,\dots,e_n\}$ (e.g. ``Washington" could mean \textbf{Washington, D.C.} or \textbf{George Washington}). Disambiguation is finding for each mention $m$ the a ground truth entity $e^{\mathrm{GT}}$ in $\mathcal{E}_m$. Typically, disambiguation operates according to two criteria: in a large corpus, how often does a mention point to an entity, $\mathrm{LinkCount}(m, e)$, and how often does entity $e_1$ co-occur with entity $e_2$, an $O(N^2)$ process, often named {\em coherence} \cite{milne2008learning,tagme,yamada2016joint}.

\subsection{Entity Linking with Types}
In this work we extend the EL task to associate with each entity a series of types (e.g. \texttt{Person}, \texttt{Place}, etc.) that if known, would rule out invalid answers, and therefore ease linking (e.g. the context now enables types to disambiguate ``Washington"). Knowledge of the types $T$ associated with a mention can also help prune entities from the the proposal set, to produce a constrained set: $\mathcal{E}_{m, T}~\subseteq~\mathcal{E}_m$.
In a probabilistic setting it is also possible to rank an entity $e$ in document $x$ according to its likelihood under the type system prediction and under the entity model:
\begin{align}
\mathbb P(e|x) \propto \mathbb P_{\mathrm{type}}(\mathrm{types}(e)|x) \cdot \mathbb P_{\mathrm{entity}}(e|x, \mathrm{types}(e)).
\end{align}
In prior work, the 112 FIGER Types \cite{ling2012fine} were associated with entities to combine an NER tagger with an EL system \cite{ling2015design}. In their work, they found that regular NER types were unhelpful, while finer grain FIGER types improved system performance.

\section{DeepType for Entity Linking}
\begin{figure*}[ht]
    \centering
    \includegraphics[width=0.85\textwidth]{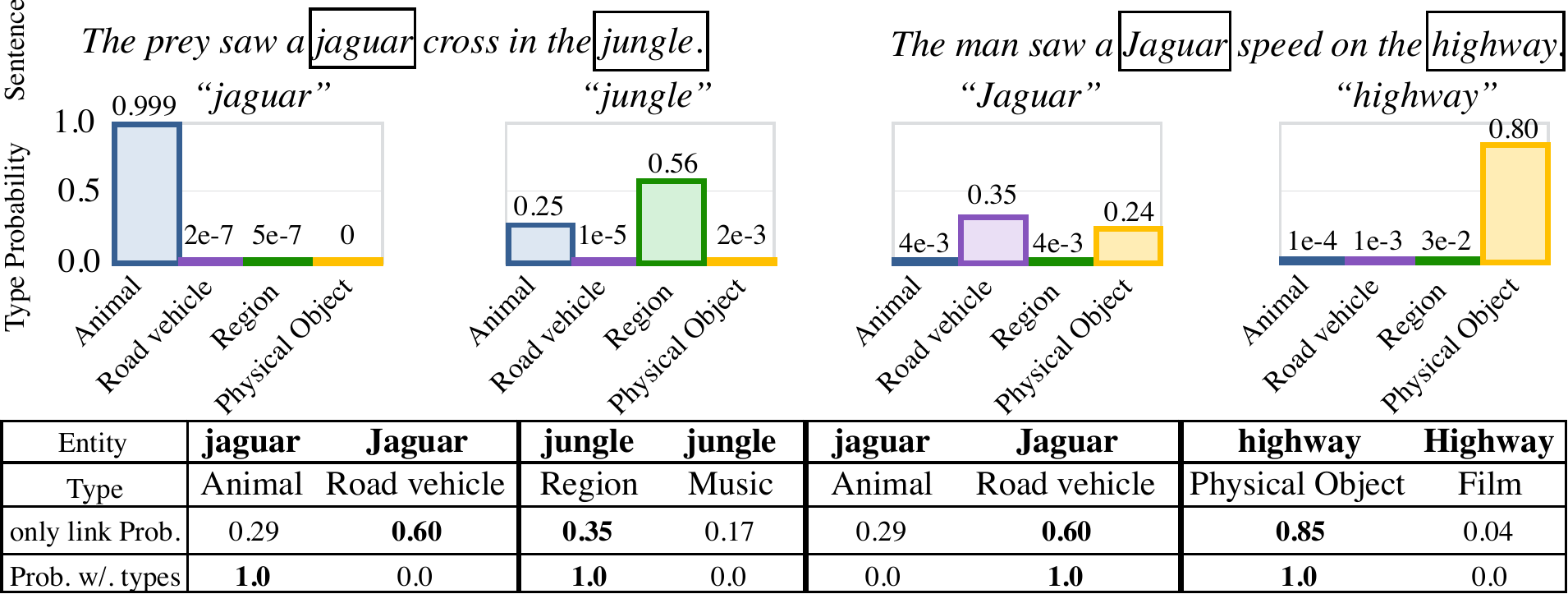}
    \caption{Example model output: ``jaguar" refers to different entities depending on context. Predicting the type associated with each word (e.g. animal, region, etc.) helps eliminate options that do not match, and recover the true entity. Bar charts give the system's belief over the type-axis ``\texttt{IsA}", and the table shows how types affects entity probabilities given by Wikipedia links.}
    \label{fig:example}
\end{figure*}

DeepType is a strategy for integrating symbolic knowledge into the reasoning process of a neural network through a type system. When we apply this technique to EL, we constrain the behavior of an entity prediction model to respect the symbolic structure defined by types. As an example, when we attempt to disambiguate ``Jaguar" the benefits of this approach are apparent: our decision can be based on whether the predicted type is Animal or Road Vehicle as shown visually in Figure~\ref{fig:example}.

In this section, we will first define key terminology, then explain the model and its sub-components separately.
\subsection{Terminology}

\begin{figure}[ht]
\centering
\includegraphics[width=0.8\linewidth]{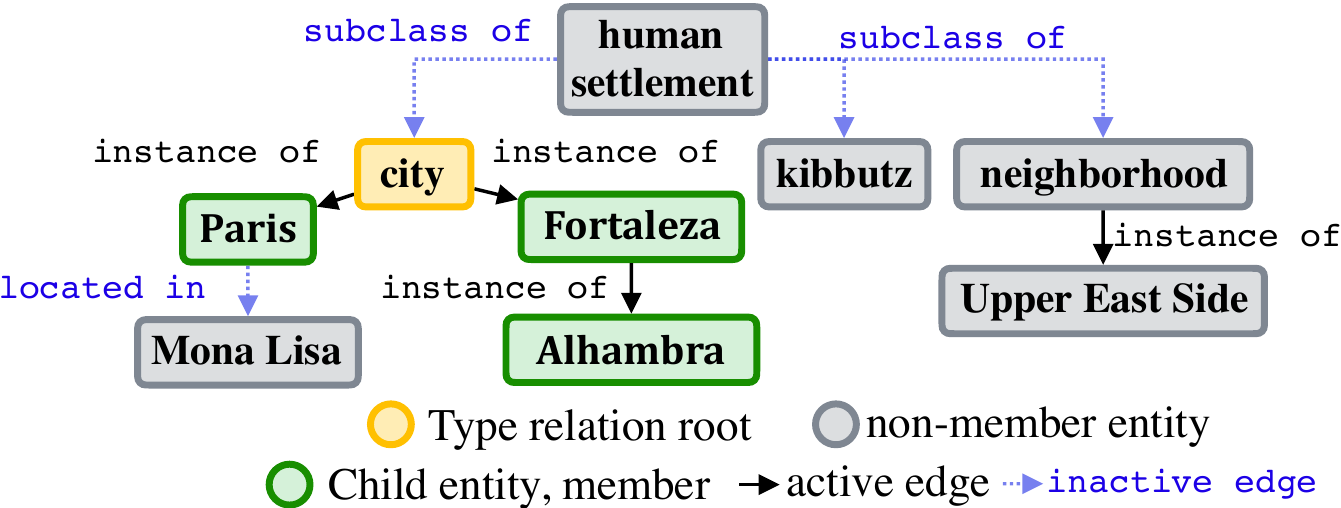}
\caption{Defining group membership with a knowledge graph relation: children of root (city) via edge (instance of).}
\label{fig:propagation}
\end{figure}

\subsubsection{Relation} Given some knowledge graph or feature set, a {\em relation} is a set of inheritance rules that define membership or exclusion from a particular group. For instance the relation {\tt instance of(}{\bf city}{\tt)} selects all children of the {\em root} {\bf city} connected by {\tt instance of} as members of the group, depicted by outlined boxes in Figure \ref{fig:propagation}.

\subsubsection{Type} In this work a {\em type} is a label defined by a {\em relation} (e.g. {\tt IsHuman}  is the type applied to all children of {\bf Human} connected by {\tt instance of}).

\subsubsection{Type Axis} A set of mutually exclusive types (e.g. $\texttt{IsHuman} \land \texttt{IsPlant} = \{\}$).

\subsubsection{Type System} A grouping of type axes, $\mathcal{A}$, along with a type labelling function: $\{t_1, \dots, t_k\} = \mathrm{TypeLabeler}(e, \mathcal{A})$. For instance a type system with two axes \{\texttt{IsA}, \texttt{Topic}\} assigns to \textbf{George Washington}: \{\texttt{Person}, \texttt{Politics}\}, and to \textbf{Washington, D.C.}: \{\texttt{Place},  \texttt{Geography}\}).

\subsection{Model}


To construct an EL system that uses type constraints we require: a type system, the associated type classifier, and a model for predicting and ranking entities given a mention.
Instead of assuming we receive a type system, classifier, entity prediction model, we will instead create the type system and its classifier starting from a given entity prediction model and ontology with text snippets containing entity mentions (e.g. Wikidata and Wikipedia). For simplicity we use $\mathrm{LinkCount}(e, m)$ as our entity prediction model.

We restrict the types in our type systems to use a set of parent-child relations over the ontology in Wikipedia and Wikidata, where each type axis has a root node $r$ and an edge type $g$, that sets membership or exclusion from the axis (e.g. $r=\textbf{human}, e=\texttt{instance of}$, splits entities into: human vs. non-human\footnote{Type ``\texttt{instance of}:\textbf{human}" mimics the NER  \texttt{PER} label.}).

We then reformulate the problem into a mixed integer problem, where discrete variables control which roots $r_1,\dots,r_k$ and edge types  $g_1,\dots,g_k$ among all roots $\mathcal{R}$ and edge types $\mathcal{G}$ will define type axes, while the continuous variables $\theta$ parametrize a classifier fit to the type system.
Our goal in type system design is to select parent-child relations that a classifier easily predicts, and where the types improve disambiguation accuracy.

\subsection{Objective}
To formally define our mixed integer problem, let us first denote $\mathcal{A}$ as the assignment for the discrete variables that define our type system (i.e. boolean variables defining if a parent-child relation gets included in our type system), $\theta$ as the parameters for our entity prediction model and type classifier, and $S_{\mathrm{model}}(\mathcal{A}, \theta)$ as the disambiguation accuracy given a test corpus containing mentions $M=\left\{(m_0, e_{0}^{\mathrm{GT}}, \mathcal{E}_{m_0}),\dots,(m_n, e_{n}^{\mathrm{GT}}, \mathcal{E}_{m_n})\right\}$. We now assume our model produces some score for each proposed entity $e$ given a mention $m$ in a document $D$, defined $\mathrm{EntityScore}(e,m,D, \mathcal{A}, \theta)$. The predicted entity for a given mention is thus: $e^*=\argmax_{e\in \mathcal{E}_m}{\mathrm{EntityScore}(e,m,D, \mathcal{A}, \theta)}$. If $e^* = e^{\mathrm{GT}}$, the mention is disambiguated. Our problem is thus defined as:
\begin{align}
\max\limits_{\mathcal{A}} \max\limits_{\theta}\,S_{\mathrm{model}}(\mathcal{A}, \theta) &= \frac{\sum\limits_{(m, e_{\mathrm{GT}}, \mathcal{E}_m)\in M} \mathbbm{1}_{e_{\mathrm{GT}}} (e^*)}{|M|}.
\end{align}


This original formulation cannot be solved exactly\footnote{There are $\sim2^{2.4\cdot 10^7}$ choices if each Wikipedia article can be a type within our type system.}. To make this problem tractable we propose a 2-step algorithm:

\begin{enumerate}
\item {\bf Discrete Optimization of Type System}: Heuristic search or stochastic optimization over the discrete variables of the type system, $\mathcal{A}$, informed by a Learnability heuristic and an Oracle.
\item {\bf Type classifier}: Gradient descent over continuous variables $\theta$ to fit type classifier and entity prediction model.
\end{enumerate}

We will now explain in more detail discrete optimization of a type system, our heuristics (Oracle and Learnability heuristic), the type classifier, and inference in this model.

\subsection{Discrete Optimization of a Type System}
\label{para:objective}

The original objective $S_{\mathrm{model}}(\mathcal{A}, \theta)$ cannot be solved exactly, thus we rely on heuristic search or stochastic optimization to find suitable assignments for $\mathcal{A}$. To avoid training an entire type classifier and entity prediction model for each evaluation of the objective function, we instead use a proxy objective $J$ for the discrete optimization\footnote{Training of the type classifier takes $\sim$3 days on a Titan X Pascal, while our Oracle can run over the test set in 100ms.}. To ensure that maximizing $J(\mathcal{A})$ also maximizes $S_{\mathrm{model}}(\mathcal{A}, \theta)$, we introduce a Learnability heuristic and an Oracle that quantify the disambiguation power of a proposed type system, an estimate of how learnable the type axes in the selected solution will be. We measure an upper bound for the disambiguation power by measuring disambiguation accuracy $S_{\mathrm{oracle}}$ for a type classifier Oracle over a test corpus.

To ensure that the additional disambiguation power of a solution $\mathcal{A}$ translates in practice we weigh by an estimate of solution's learnability $\mathrm{Learnability}(\mathcal{A})$ improvements between $S_{\mathrm{oracle}}$ and the accuracy of a system that predicts only according to the entity prediction model\footnote{For an entity prediction model based only on link counts, this means always picking the most linked entity.}, $S_{\mathrm{greedy}}$.

Selecting a large number of type axes will provide strong disambiguation power, but may lead to degenerate solutions that are harder to train, slow down inference, and lack higher-level concepts that provide similar accuracy with less axes. We prevent this by adding a per type axis penalty of $\lambda$.

Combining these three terms gives us the equation for $J$:
\begin{equation}
\begin{split}
\label{eq:obj}
J(\mathcal{A}) =& (S_{\mathrm{oracle}} - S_{\mathrm{greedy}}) \cdot \mathrm{Learnability}(\mathcal{A})\,+ \\
&S_{\mathrm{greedy}} - |\mathcal A| \cdot \lambda
.
\end{split}
\end{equation}
\subsubsection{Oracle}
\label{para:oracle}
Our Oracle is a methodology for abstracting away machine learning performance from the underlying representational power of a type system $\mathcal{A}$. It operates on a test corpus with a set of mentions, entities, and proposal sets: $m_i, e_{i}^{\mathrm{GT}}, \mathcal{E}_{m_i}$. The Oracle prunes each proposal set to only contain entities whose types match those of $e_{i}^{\mathrm{GT}}$, yielding $\mathcal{E}_{m,\mathrm{oracle}}$. Types fully disambiguate when $|\mathcal{E}_{m,\mathrm{oracle}}|=1$, otherwise we use the entity prediction model to select the right entity in the remainder set $\mathcal{E}_{m_i,\mathrm{oracle}}$:
\begin{align}
 \mathrm{Oracle}(m) = \argmax_{e \in \mathcal{E}_{m, \mathrm{oracle}}}\mathbb P_{\mathrm{entity}}(e|m, \mathrm{types}(x)).
\end{align}
If $\mathrm{Oracle}(m) = e^{\mathrm{GT}}$, the mention is disambiguated. Oracle accuracy is denoted $S_{\mathrm{oracle}}$ given a type system over a test corpus containing mentions $M=\left\{(m_0, e_{0}^{\mathrm{GT}}, \mathcal{E}_{m_0}),\dots,(m_n, e_{n}^{\mathrm{GT}}, \mathcal{E}_{m_n})\right\}$:
\begin{align}
S_{\mathrm{oracle}} &= \frac{\sum_{(m, e_{\mathrm{GT}}, \mathcal{E}_m)\in M} \mathbbm{1}_{e_{\mathrm{GT}}} (\mathrm{Oracle}(m))}{|M|}.
\end{align}

\paragraph{Learnability}
\begin{figure}[ht]
    \centering
    \begin{subfigure}{0.45\linewidth}
    \includegraphics[width=\linewidth]{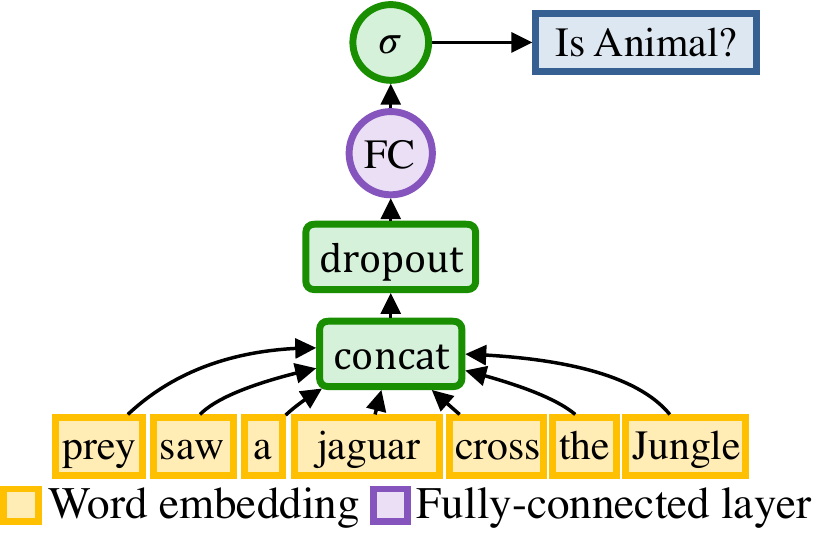}
    \caption{}
    \label{fig:windowclassifier}
    \end{subfigure}
    \begin{subfigure}{0.45\linewidth}
    \includegraphics[width=1.3in]{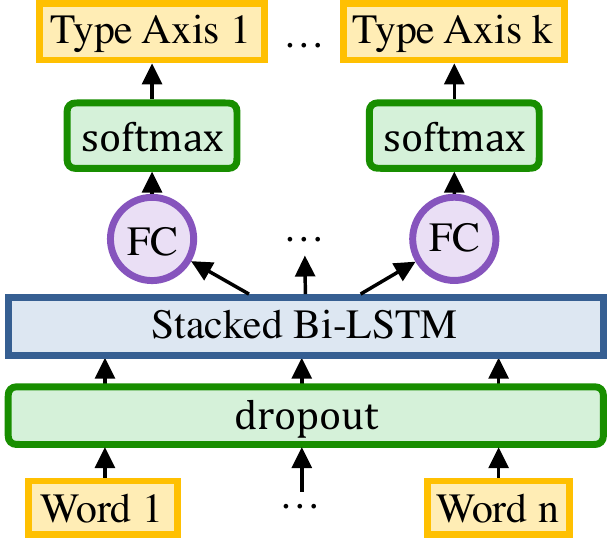}
    \caption{}
    \label{fig:bilstm}
    \end{subfigure}
    \caption{Text window classifier in (a) serves as type Learnability estimator, while the network in (b) takes longer to train, but discovers long-term dependencies to predict types and jointly produces a distribution for multiple type axes.}
\end{figure}

To ensure that disambiguation gains obtained during the discrete optimization are available when we train our type classifier, we want to ensure that the types selected are easy to predict. The Learnability heuristic empirically measures the average performance of classifiers at predicting the presence of a type within some Learnability-specific training set.

To efficiently estimate Learnability for a full type system we make an independence assumption and model it as the mean of the Learnability for each individual axis, ignoring positive or negative transfer effects between different type axes. This assumption lets us parallelize training of simpler classifiers for each type axis. We measure the area under its receiver operating characteristics curve (AUC) for each classifier and compute the type system's learnability: $\mathrm{Learnability}(\mathcal{A}) = \frac{\sum_{t \in \mathcal{A}}{\mathrm{AUC}(t)}}{|\mathcal{A}|}$.
We use a text window classifier trained over windows of 10 words before and after a mention. Words are represented with randomly initialized word embeddings; the classifier is illustrated in Figure \ref{fig:windowclassifier}. AUC is averaged over 4 training runs for each type axis.

\subsection{Type Classifier}
\label{section:typeclassifier}

After the discrete optimization has completed we now have a type system  $\mathcal{A}$. We can now use this type system to label data in multiple languages from text snippets associated with the ontology\footnote{Wikidata's ontology has cross-links with Wikipedia, IMDB, Discogs, MusicBrainz, and other encyclopaedias with snippets.}, and supervize a Type classifier.

The goal for this classifier is to discover long-term dependencies in the input data that let it reliably predict types across many contexts and languages. For this reason we select a bidirectional-LSTM \cite{lample2016neural} with word, prefix, and suffix embeddings as done in \cite{andor2016globally}. Our network is shown pictorially in Figure \ref{fig:bilstm}.
Our classifier is trained to minimize the negative log likelihood of the per-token types for each type axis in the document $D$ with $L$ tokens: $-\sum_{i=1}^k\log \mathbb P_i(t_{i,1}, \dots, t_{i,L}| D)$. When using Wikipedia as our source of text snippets our label supervision is partial\footnote{We obtain type labels only on the intra-wiki link anchor text.}, so we make a conditional independence assumption about our predictions and use Softmax as our output activation:
$-\sum_{i=1}^k\sum_{j=1}^L\log \mathbb P_i(t_{i,j}| w_j, D)$.

\subsection{Inference}
\label{para:inference-el}
At inference-time we incorporate classifier belief into our decision process by first running it over the full context and obtaining a belief over each type axis for each input word $w_0,\dots,w_L$. For each mention $m$ covering words $w_x,\dots,w_y$, we obtain the type conditional probability for all type axes $i$: $\left\{\mathbb P_i(\cdot|w_x,D),\dots,\mathbb P_i(\cdot|w_y,D)\right\}$. In multi-word mentions we must combine beliefs over multiple tokens $x\dots y$: the product of the beliefs over the mention's tokens is correct but numerically unstable and slightly less performant than max-over-time\footnote{The choice of max-over-time is empirically motivated: we compared product mean, min, max, and found that max was comparable to mean, and slightly better than the alternatives.}, which we denote for the $i$-th type axis: $\mathbb P_{i,*}(\cdot|m,D)$.

The score  $s_{e,m,D, \mathcal{A}, \theta} = \mathrm{EntityScore}(e,m,D, \mathcal{A}, \theta)$ of an entity $e$ given these conditional probability distributions $\mathbb P_{1,*}(\cdot|m,D),\dots,\mathbb P_{k,*}(\cdot|m,D)$, and the entities' types in each axis $t_1,\dots,t_k$ can then be combined to rank entities according to how predicted they were by both the entity prediction model and the type system. The chosen entity $e^*$ for a mention $m$ is chosen by taking the option that maximizes the score among the $\mathcal{E}_m$ possible entities; the equation for scoring and $e^*$ is given below, with $\mathbb P_{\mathrm{Link}}(e | m) = \frac{ \mathrm{LinkCount}(m, e) }{ \sum_{j \in \mathcal{E}_{m}}\mathrm{LinkCount}(m, j)}$, $\alpha_i$ a per type axis smoothing parameter, $\beta$ is a smoothing parameter over all types:

\begin{equation}
\begin{split}
s_{e,m,D, \mathcal{A}, \theta} =&\mathbb P_{\mathrm{Link}}(e | m) \cdot \Big(1 - \beta + \beta\,\cdot\\
  &\left\{\prod_{i=1}^k (1 - \alpha_i + \alpha_i \cdot \mathbb P_{i,*}(t_i|m,D)) \right\}\Big).
\end{split}
\end{equation}

\section{Results}

\subsection{Type System Discovery}
In the following experiments we evaluate the behavior of different search methodologies for type system discovery: which method best scales to large numbers of types, achieves high accuracy on the target EL task, and whether the choice of search impacts learnability by a classifier or generalisability to held-out EL datasets.

For the following experiments we optimize DeepType's type system over a held-out set of 1000 randomly sampled articles taken from the Feb. 2017 English Wikipedia dump, with the Learnability heuristic text window classifiers trained only on those articles. The type classifier is trained jointly on English and French articles, totalling 800 million tokens for training, 1 million tokens for validation, sampled equally from either language.

We restrict roots $\mathcal{R}$ and edges $\mathcal{G}$ to the most common $1.5\cdot10^5$ entities that are entity parents through \texttt{wikipedia category} or \texttt{instance of} edges, and eliminate type axes where $\mathrm{Learnability}(\cdot)$ is 0, leaving 53,626 type axes. 

\subsubsection{Human Type System Baseline}
\label{section:humandesign}
To isolate discrete optimization from system performance and gain perspective on the difficulty and nature of the type system design we incorporate a human-designed type system. Human designers have access to the full set of entities and relations in Wikipedia and Wikidata, and compose different inheritance rules through Boolean algebra to obtain higher level concepts (e.g. $\texttt{woman} = \texttt{IsHuman} \land \texttt{IsFemale}$, or $\texttt{animal} = \texttt{IsTaxon} \land \neg \{\texttt{IsHuman} \lor \texttt{IsPlant}\}$\footnote{Taxon is the general parent of living items in Wikidata.}). The final human system uses 5 type axes\footnote{\texttt{IsA}, \texttt{Topic}, \texttt{Location}, \texttt{Continent}, and \texttt{Time}.}, and 1218 inheritance rules.

\subsubsection{Search methodologies}
\paragraph{Beam Search and Greedy selection} We iteratively construct a type system by choosing among all remaining type axes and evaluating whether the inclusion of a new type axis improves our objective: $J(\mathcal{A} \cup \{t_j\}) > J(\mathcal{A})$. We use a beam size of $b$ and stop the search when all solutions stop growing.
\paragraph{Cross-Entropy Method} 
(CEM) \cite{rubinstein1999cross} is a stochastic optimization procedure applicable to the selection of types. We begin with a probability vector $\vec{P_0}$ set to $p_{\mathrm{start}}$, and at each iteration we sample $M_{\mathrm{CEM}}$ vectors $\vec{s}$ from the Bernoulli distribution given by $\vec{P_i}$, and measure each sample's fitness with Eq.~\ref{eq:obj}. The $N_{\mathrm{CEM}}$ highest fitness elements are our winning population $\mathcal{S}_t$ at iteration $t$. Our probabilities are fit to $\mathcal{S}_t$ giving $P_{t+1} = \frac{\sum_{\vec{s} \in \mathcal{S}_t}\vec{s}}{N_{\mathrm{CEM}}}$. The optimization is complete when the probability vector is binary.

\paragraph{Genetic Algorithm} The best subset of type axes can be found by representing type axes as genes carried by $N_{\mathrm{population}}$ individuals in a population undergoing mutations and crossovers \cite{harvey2009microbial} over $G$ generations. We select individuals using Eq.~\ref{eq:obj} as our fitness function.
\subsubsection{Search Methodology Performance Impact}
\begin{table*}[ht]
\begin{center}
\caption{Method comparisons. Highest value in {\bf bold}, excluding oracles.}
    \begin{subtable}{.42\linewidth}
\caption{Type system discovery method comparison}
\begin{center}
\begin{tabular}{ |l|r|r|r|}
\hline
Approach        & Evals  & Accuracy & Items\\
\hline
BeamSearch & $5.12\cdot 10^7$ & 97.84  & 130\\
Greedy     & $6.40\cdot 10^6$ & 97.83  & 130\\
GA          & $116,000$        & 96.959 & 128\\
CEM        & $43,000$         & 96.26  & 89\\
Random     & N/A & $92.9\pm0.28$ & 128 \\
No types    & 0 & $92.10 $ & 0 \\
\hline
\end{tabular}
\end{center}
\label{table:generations}
    \end{subtable}%
    \begin{subtable}{.55\linewidth}
\caption{NER F1 score comparison for DeepType pretraining vs. baselines.}
\begin{center}
\begin{tabular}{ |l|r|r|r|r|}
\hline
     \multirow{2}{*}{Model} & \multicolumn{2}{|c|}{CoNLL 2003} & \multicolumn{2}{|c|}{OntoNotes}\\
\cline{2-5}
& Dev & Test & Dev & Test \\
\hline
Bi-LSTM  & \multirow{2}{*}{-}& \multirow{2}{*}{76.29} &  \multirow{2}{*}{-} & \multirow{2}{*}{77.77} \\
\cite{chiu2015named} & & & & \\
Bi-LSTM-CNN + emb + lex & \multirow{2}{*}{\bf 94.31} & \multirow{2}{*}{\bf 91.62} & \multirow{2}{*}{84.57} & \multirow{2}{*}{\bf 86.28} \\
\cite{chiu2015named} & & & & \\
\hline
Bi-LSTM (Ours)                      & 89.49 & 83.40 & 82.75 & 81.03\\
Bi-LSTM-CNN (Ours)                  & 90.54 & 84.74 & 83.17 & 82.35\\
Bi-LSTM-CNN (Ours) + types          & 93.54 & 88.67 & {\bf 85.11} & 83.12\\
\hline
\end{tabular}
\end{center}
\label{table:ner}
    \end{subtable}
    \begin{subtable}{1\linewidth}
    \caption{Entity Linking model Comparison. Significant improvements over prior work denoted by \oneS\, for $p<0.05$, and \twoS\, for $p<0.01$.}
\begin{center}
\begin{tabular}{ |c|r|l|l|l|l|l|l|l|}
\hline
     \multicolumn{2}{|l|}{Model} & enwiki & frwiki & dewiki & eswiki & \textsc{WKD30} & CoNLL & TAC 2010\\
\hline
\multicolumn{2}{|l|}{M\&W\cite{milne2008learning}} &        & & & & 84.6  & - & -\\
\multicolumn{2}{|l|}{TagMe \cite{tagme}}                       & 83.224 & & 80.711 & & 90.9 & - & -\\
\multicolumn{2}{|l|}{\cite{globerson2016collective}}      &
   & & & &        - & 91.7 & 87.2\\
\multicolumn{2}{|l|}{\cite{yamada2016joint}}      &        & & & &   - & 91.5 & 85.2\\
\multicolumn{2}{|l|}{NTEE \cite{yamada2017learning}}      &        & & & &        - & - & 87.7\\
\hline
\multicolumn{2}{|l|}{$\mathrm{LinkCount}$ only }    & 89.064\twoS
&  92.013 &  92.013\twoS &  89.980 & 82.710 &  68.614 & 81.485\\
\hline
\multirow{10}{*}{\rot{Ours}} & manual                & {\bf94.331\twoS} & 92.967 & &        & 91.888\twoS & 93.108\twoS & 90.743\oneS \\
& manual (oracle)                                    & 97.734  & 98.026 & 98.632 & 98.178 & 95.872 & 98.217 & 98.601\\
& greedy                                             & 93.725\twoS  & {\bf 92.984} & & & {\bf 92.375\twoS} & 94.151\twoS & {\bf90.850\oneS}\\
& greedy (oracle)                                    & 98.002  & 97.222 & 97.915 & 98.246 & 97.293 & 98.982 & 98.278\\
& CEM                                                & 93.707\twoS  & 92.415 & & & 92.247\twoS & 93.962\twoS & 90.302\oneS\\
& CEM (oracle)                                       & 97.500  & 96.648 & 97.480 & 97.599 & 96.481 & 99.005 & 96.767\\
& GA                                                 & 93.684\twoS  & 92.027 & & & 92.062\twoS & {\bf 94.879\twoS} & 90.312\oneS\\
& GA (oracle)                                        & 97.297  & 96.783 & 97.408 & 97.609 & 96.268 & 98.461 & 96.663\\
& GA (English only)                                  & 93.029\twoS & & & & 91.743\twoS & 93.701\twoS & - \\
\hline
\end{tabular}
\end{center}
\label{table:comparison}
    \end{subtable}
    \end{center}
\end{table*}

To validate that $\lambda$ controls type system size, and find the best tradeoff between size and accuracy, we experiment with a range of values and find that accuracy grows more slowly below 0.00007, while system size still increases.

From this point on we keep $\lambda=0.00007$, and we compare the number of iterations needed by different search methods to converge, against two baselines: the empty set and the mean performance of 100 randomly sampled sets of 128 types (Table \ref{table:generations}). We observe that the performance of stochastic optimizers GA and CEM is similar to heuristic search, but requires orders of magnitude less function evaluations.

Next, we compare the behavior of the different search methods to a human designed system and state of the art approaches on three standard datasets (i.e. \textsc{Wiki-Disamb30} (WKD30) \cite{tagme}\footnote{We apply the preprocessing and link pruning as  \cite{tagme} to ensure the comparison is fair.}, CoNLL(YAGO) \cite{AIDA2011}, and TAC KBP 2010 \cite{ji2010overview}), along with test sets built by randomly sampling 1000 articles from Wikipedia's February 2017 dump in English, French, German, and Spanish which were excluded from training the classifiers. Table \ref{table:comparison} has Oracle performance for the different search methods on the test sets, where we report disambiguation accuracy per annotation. A $\mathrm{LinkCount}$ baseline is included that selects the mention's most frequently linked entity\footnote{Note that LinkCount accuracy is stronger than the one found in \cite{tagme} or \cite{milne2008learning} because newer Wikipedia dumps improve link coverage and reduce link distribution noisiness.}. All search techniques' Oracle accuracy significantly improve over $\mathrm{LinkCount}$, and achieve near perfect accuracy on all datasets (97-99\%); furthermore we notice that performance between the held-out Wikipedia sets and standard datasets sets is similar, supporting the claim that the discovered type systems generalize well. We note that machine discovered type systems outperform human designed systems: CEM beats the human type system on English Wikipedia, and all search method's type systems outperform human systems on \textsc{Wiki-Disamb30}, CoNLL(YAGO), and TAC KBP 2010.

\subsubsection{Search Methodology Learnability Impact}
To understand whether the type systems produced by different search methods can be trained similarly well we compare the type system built by GA, CEM, greedy, and the one constructed manually. EL Disambiguation accuracy is shown in Table \ref{table:comparison}, where we compare with recent deep-learning based approaches \cite{globerson2016collective}, or recent work by Yamada et al. for embedding word and entities \cite{yamada2016joint}, or documents and entities \cite{yamada2017learning}, along with count and coherence based techniques Tagme \cite{tagme} and Milne \& Witten \cite{milne2008learning}. To obtain Tagme's Feb. 2017 Wikipedia accuracy we query the public web API\footnote{\url{https://tagme.d4science.org/tagme/}} available in German and English, while other methods can be compared on CoNLL(YAGO) and TAC KBP 2010. Models trained on a human type system outperform all previous approaches to entity linking, while type systems discovered by machines lead to even higher performance on all datasets except English Wikipedia.

\subsection{Cross-Lingual Transfer}
Type systems are defined over Wikidata/Wikipedia, a multi-lingual knowledge base/encyclopaedia, thus type axes are language independent and can produce cross-lingual supervision. To verify whether this cross-lingual ability is useful we train a type system on an English dataset and verify whether it can successfully supervize French data. We also measure using the Oracle (performance upper bound) whether the type system is useful in Spanish or German. Oracle performance across multiple languages does not appear to degrade when transferring to other languages (Table \ref{table:comparison}). We also notice that training in French with an English type system still yields improvements over $\mathrm{LinkCount}$ for CEM, greedy, and human systems.

Because multi-lingual training might oversubscribe the model, we verified if monolingual would outperform bilingual training: we compare GA in English + French with only English (last row of Table \ref{table:comparison}). Bilingual training does not appear to hurt, and might in fact be helpful.

We follow-up by inspecting whether the bilingual word vector space led to shared representations: common nouns have their English-French translation close-by, while proper nouns do not (French and US politicians cluster separately).
\subsection{Named Entity Recognition Transfer}
The goal of our NER experiment is to verify whether DeepType produces a type sensitive language representation useful for transfer to other downstream tasks. To measure this we pre-train a type classifier with a character-CNN and word embeddings as inputs, following \cite{kim2015character}, and replace the output layer with a linear-chain CRF \cite{lample2016neural} to fine-tune to NER data. Our model's F1 scores when transferring to the CoNLL 2003 NER task and OntoNotes 5.0 (CoNLL 2012) split are given in Table \ref{table:ner}. We compare with two baselines that share the architecture but are not pre-trained, along with the current state of the art \cite{chiu2015named}.

We see positive transfer on Ontonotes and CoNLL: our baseline Bi-LSTM strongly outperforms \cite{chiu2015named}'s baseline, while pre-training gives an additional 3-4 F1 points, with our best model outperforming the state of the art on the OntoNotes development split. While our baseline LSTM-CRF performs better than in the literature, our strongest baseline (CNN+LSTM+CRF) does not match the state of the art with a lexicon. We find that DeepType always improves over baselines and partially recovers lexicon performance gains, but does not fully replace lexicons.

\section{Related Work}
\paragraph{Neural Network Reasoning with Symbolic structures} Several approaches exist for incorporating symbolic structures into the reasoning process of a neural network by designing a loss function that is defined with a label hierarchy. In particular the work of \cite{deng2012hedging} trades off specificity for accuracy, by leveraging the hyper/hyponymy relation to make a model aware of different granularity levels. Our work differs from this approach in that we design our type system within an ontology to meet specific accuracy goals, while they make the accuracy/specificity tradeoff at training time, with a fixed structure. More recently \cite{wu2017hierarchical} use a hierarchical loss to increase the penalty for distant branches of a label hierarchy using the ultrametric tree distance. We also aim to capture the most important aspects of the symbolic structure and shape our loss function accordingly, however our loss shaping is a result of discrete optimization and incorporates a learnability heuristic to choose aspects that can easily be acquired.

A different direction for integrating structure stems from constraining model outputs, or enforcing a grammar. In the work of \cite{ling2015design}, the authors use NER and FIGER types to ensure that an EL model follows the constraints given by types. We also use a type system and constrain our model's output, however our type system is task-specific and designed by a machine with a disambiguation accuracy objective, and unlike the authors we find that types improve accuracy. The work of \cite{jayant2017parsing} uses a type-aware grammar to constrain the decoding of a neural semantic parser. Our work makes use of type constraints during decoding, however the grammar and types in their system require human engineering to fit each individual semantic parsing task, while our type systems are based on online encyclopaedias and ontologies, with applications beyond EL.

\paragraph{Neural Entity Linking} Current approaches to entity linking make extensive use of deep neural networks, distributed representations. In \cite{globerson2016collective} a neural network uses attention to focus on contextual entities to disambiguate. While our work does not make use of attention, RNNs allow context information to affect disambiguation decisions. In the work of \cite{yamada2016joint} and \cite{yamada2017learning}, the authors adopt a distributed representation of context which either models words and entities, or documents and entities such that distances between vectors informs disambiguation. We also rely on word and document vectors produced by RNNs, however entities are not explicitly represented in our neural network, and we use context to predict entity types, thereby allowing us to incorporate new entities without retraining.
\section{Conclusion}

In this work we introduce DeepType, a method for integrating symbolic knowledge into the reasoning process of a neural network. We've proposed a mixed integer reformulation for jointly designing type systems and training a classifier for a target task, and empirically validated that when this technique is applied to EL it is effective at integrating symbolic information in the neural network reasoning process.
When pre-training with DeepType for NER, we observe improved performance over baselines and a new state of the art on the OntoNotes dev set, suggesting there is cross-domain transfer: symbolic information is incorporated in the neural network's distributed representation.
Furthermore we find that type systems designed by machines outperform those designed by humans on three benchmark datasets, which is attributable to incorporating learnability and target task performance goals within the design process.
Our approach naturally enables multilingual training, and our experiments show that bilingual training improves over monolingual, and type systems optimized for English operate at similar accuracies in French, German, and Spanish, supporting the claim that the type system optimization leads to the discovery of high level cross-lingual concepts useful for knowledge representation.
We compare different search techniques, and observe that stochastic optimization has comparable performance to heuristic search, but with orders of magnitude less objective function evaluations.

The main contributions of this work are a joint formulation for designing and integrating symbolic information into neural networks, that enable us to constrain the outputs to obey symbolic structure, and an approach to EL that uses type constraints. Our approach reduces EL resolution complexity from $O(N^2)$ to $O(N)$, while allowing new entities to be incorporated without retraining, and we find on three standard datasets (WikiDisamb30, CoNLL (YAGO), TAC KBP 2010) that our approach outperforms all existing solutions by a wide margin, including approaches that rely on a human-designed type system \cite{ling2015design} and the more recent work by Yamada et al. for embedding words and entities \cite{yamada2016joint}, or document and entities \cite{yamada2017learning}. As a result of our experiments, we observe that disambiguation accuracy using Oracles reaches 99.0\% on CoNLL (YAGO) and 98.6\% on TAC KBP 2010, suggesting that EL would be almost solved if we can close the gap between type classifiers and the Oracle.

The results presented in this work suggest many directions for future research: we may test how DeepType can be applied to other problems where incorporating symbolic structure is beneficial, whether making type system design more expressive by allowing hierarchies can help close the gap between model and Oracle accuracy, and seeing if additional gains can be obtained by relaxing the classifier's conditional independence assumption.
\subsubsection*{Acknowledgments}
We would like to thank the anonymous reviewers for their valuable feedback. In addition, we thank John Miller, Andrew Gibiansky, and Szymon Sidor for thoughtful comments and fruitful discussion.
\normalsize
\appendix
\appendixpage
\section{Training details and hyperparameters}
\label{section:hyperparameters}
\subsection{Optimization}

Our models are implemented in Tensorflow and optimized with Adam with a learning rate of $10^{-4}$, $\beta_1=0.9, \beta_2=0.999, \epsilon=10^{-8}$, annealed by 0.99 every 10,000 iterations. 

To reduce over-fitting and make our system more robust to spelling changes we apply Dropout to input embeddings and augment our data with noise: swap input words with a special \texttt{<UNK>} word, remove capitalization or a trailing ``s." In our NER experiments we add Gaussian noise during training to the LSTM weights with $\sigma=10^{-6}$.

We use early stopping in our NER experiments when validation F1 score stops increasing. Type classification model selection is different as the models did not overfit, thus we instead stop training when no more improvements in F1 are observed on held-out type-training data ($\sim3$ days on one Titan X Pascal).
\begin{table}[ht]
\caption{Hyperparameters for type system discovery search.}
\begin{center}
\begin{tabular}{|r|r|r|}
\hline
Method & Parameter & Value\\
\hline
Greedy           & $b$ & 1\\
Beam Search & $b$ & 8\\
\hline
\multirow{3}{*}{\rot{CEM}} & $M_\mathrm{CEM}$ & 1000\\
& $p_{\mathrm{start}}$ & $\frac{50}{|\mathcal{R}|} \approx 0.001$\footnotemark\\
& $N_\mathrm{CEM}$ & 200\\
\hline
\multirow{4}{*}{\rot{GA}} & $G$ & 200\\
& $N_{\mathrm{population}}$ & 1000\\
& mutation probability & 0.5\\
& crossover probability & 0.2\\
\hline
\end{tabular}
\end{center}
\label{table:searchhyperparameters}
\end{table}
\subsection{Architecture}
\paragraph{Character representation} Our character-convolutions have character filters with (width, channels): $\{(1,50), (2, 75), (3, 75), (4, 100), (5, 200), (6, 200), (7, 200)\},$ a maximum word length of 40, and 15-dimensional character embeddings followed by 2 highway layers. We learn 6-dimensional embeddings for 2 and 3 character prefixes and suffixes.
\paragraph{Text Window Classifier} The text window classifiers have 5-dimensional word embeddings, and use Dropout of 0.5. Empirically we find that two passes through the dataset with a batch size of 128 is sufficient for the window classifiers to converge. Additionally we train multiple type axes in a single batch, reaching a training speed of 2.5 type axes/second.

\footnotetext{\label{collapse}The choice of $p_{\mathrm{start}}$ affects the system size at the first step of the CEM search: setting it too low leads to poor search space exploration, while too high increase the cost of the objective function evaluation. Empirically we know that for a given $\lambda$ the solution will have an expected size $s$. Setting $p_{\mathrm{start}}=\frac{s}{|\mathcal{R}|}$ leads to sufficient exploration to reach the performance of larger $p_{\mathrm{start}}$.}

\section{Wikipedia Link Simplification}
\label{para:anaphora}
Link statistics collected on large corpuses of entity mentions are extensively used in entity linking. These statistics provide a noisy estimate of the conditional probability of an entity $e$ for a mention $m$ $\mathbb P(e | m)$. Intra-wiki links in Wikipedia provide a multilingual and broad coverage source of links, however annotators often create link anaphoras: ``king" $\to$ \textbf{Charles I of England}. This behavior increases polysemy (``king" mention has 974 associated entities) and distorts link frequencies (``queen" links to the band \textbf{Queen} 4920 times, \textbf{Elizabeth II} 1430 times, and \textbf{monarch} only 32 times).

Problems with link sparsity or anaphora were previously identified, however present solutions rely on pruning rare links and thus lose track of the original statistics \cite{tagme,hasibi2016reproducibility,ling2015design}. We propose instead to detect anaphoras and recover the generic meaning through the Wikidata property graph: if a mention points to entities A and B, with A being more linked than B, and A is B's parent in the Wikidata property graph, then replace B with A. We define A to be the parent of B if they connect through a sequence of Wikidata properties \{{\tt instance of}, {\tt subclass of}, {\tt is a list of}\}, or through a single edge in \{{\tt occupation}, {\tt position held}, {\tt series}\footnote{e.g. \textbf{Return of the Jedi} $\underset{\text{\tt series}}{\to}$ \textbf{Star Wars}}\}. The simplification process is repeated until no more updates occur. This transformation reduces the number of associated entities for each mention (``king" senses drop from 974 to 143) and ensures that the semantics of multiple specific links are aggregated (number of ``queen" links to {\em monarch} increase from 32 to 3553).
\begin{figure}[ht]
\centering
\includegraphics[height=1.5in]{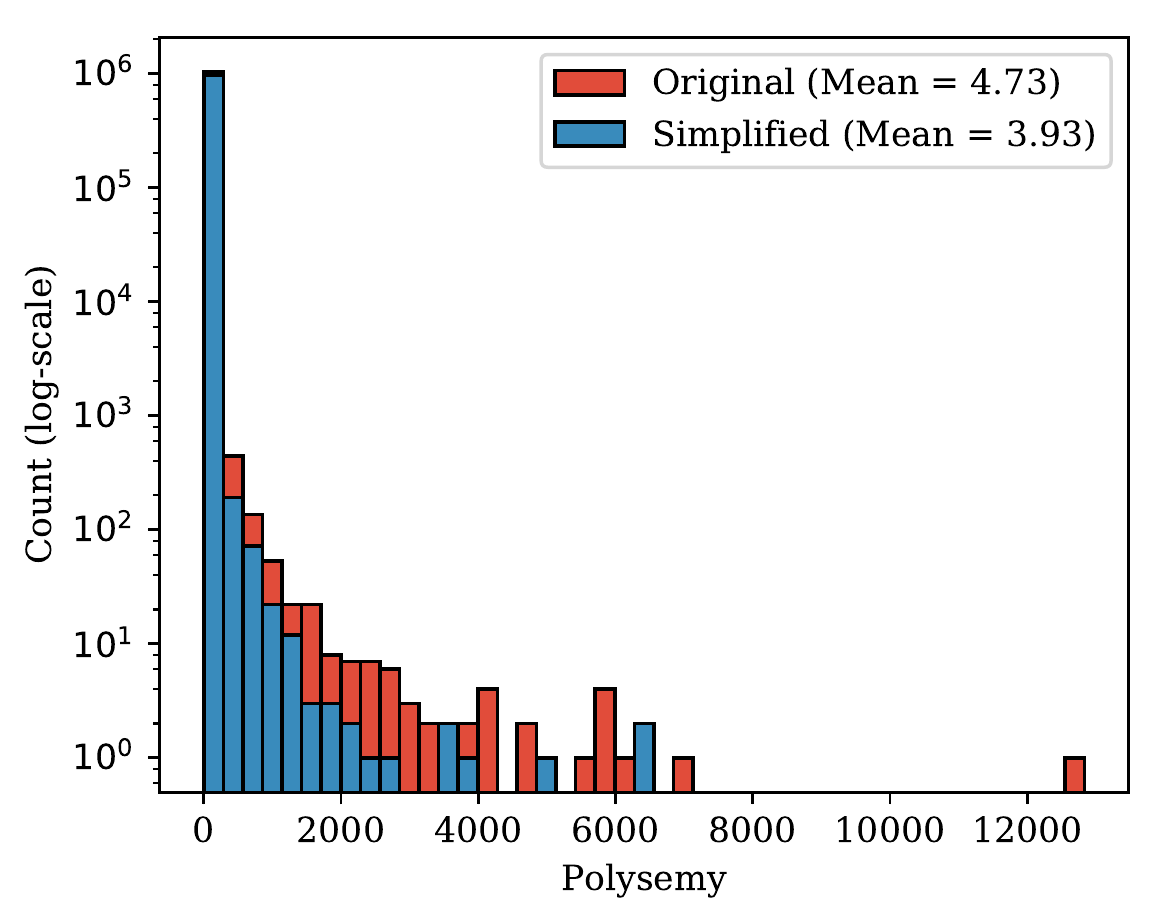}
\caption{Mention Polysemy change after simplification.}
\label{fig:polysemy}
\end{figure}

\begin{table}
\caption{Link change statistics per iteration during English Wikipedia Anaphora Simplification.}
\centering
\begin{tabular}{ |r|r|r|}
\hline
Step & Replacements & Links changed\\
\hline
1 & 1,109,408 &  9,212,321\\
2 & 13922 & 1,027,009 \\
3 & 1229 & 364,500\\
4 & 153 & 40,488\\
5 & 74 & 25,094\\
6 & 4 & 1,498\\
\hline
\end{tabular}
\label{table:polysemy}
\end{table}

After simplification we find that the mean number of senses attached to polysemous mentions drops from 4.73 to 3.93, while over 10,670,910 links undergo changes in this process (Figure \ref{fig:polysemy}). Table \ref{table:polysemy} indicates that most changes result from mentions containing entities and their immediate parents. This simplification method strongly reduces the number of entities tied to each Wikipedia mention in an automatic fashion across multiple languages.

\section{Multilingual Training Representation}
\label{section:translation}

\begin{table*}[t]
\caption{Top-$k$ Nearest neighbors (cosine distance) in shared English-French word vector space.}
\begin{center}
\begin{tabular}{|r|l|l|l|l|l|l|l|}
\hline
$k$ & Argentinian & lui & Sarkozy & Clinton & hypothesis\\
\hline
1   & argentin (0.259)   & he (0.333) & Bayron (0.395) & Reagan (0.413) & paradox (0.388)\\
2    & Argentina (0.313)   & il (0.360) & Peillon (0.409) & Trump (0.441) & Hypothesis (0.459) \\
3    & Argentine (0.315)   & him (0.398) & Montebourg (0.419) & Cheney (0.495) & hypoth\`ese (0.497) \\
\hline
\end{tabular}
\end{center}
\label{table:translate}
\end{table*}

Multilingual data creation is a side-effect of the ontology-based automatic labeling scheme. In Table \ref{table:translate} we present nearest-neighbor words for words in multiple languages. We note that common words (he, Argentinian, hypothesis) remain close to their foreign language counterpart, while proper nouns group with country/language-specific terms.
We hypothesize that common words, by not fulfilling a role as a label, can therefore operate in a language independent way to inform the context of types, while proper nouns will have different type requirements based on their labels, and thus will not converge to the same representation.

\begin{table*}[ht]
\caption{Additional set of Top-$k$ Nearest neighbors (cosine distance) in shared English-French word vector space.}
\begin{center}
\begin{tabular}{|r|l|l|}
\hline
$k$ & feu             & computer\\
\hline
1    & killing (0.585) & Computer (0.384)\\
2    & terrible (0.601) & computers (0.446) \\
3    & beings (0.618) & informatique (0.457) \\
\hline
\end{tabular}
\end{center}
\label{table:translate-more}
\end{table*}

\section{Effect of System Size Penalty}
\label{section:lambdaexp}

\begin{figure}[ht]
\centering
\includegraphics[height=1.6in]{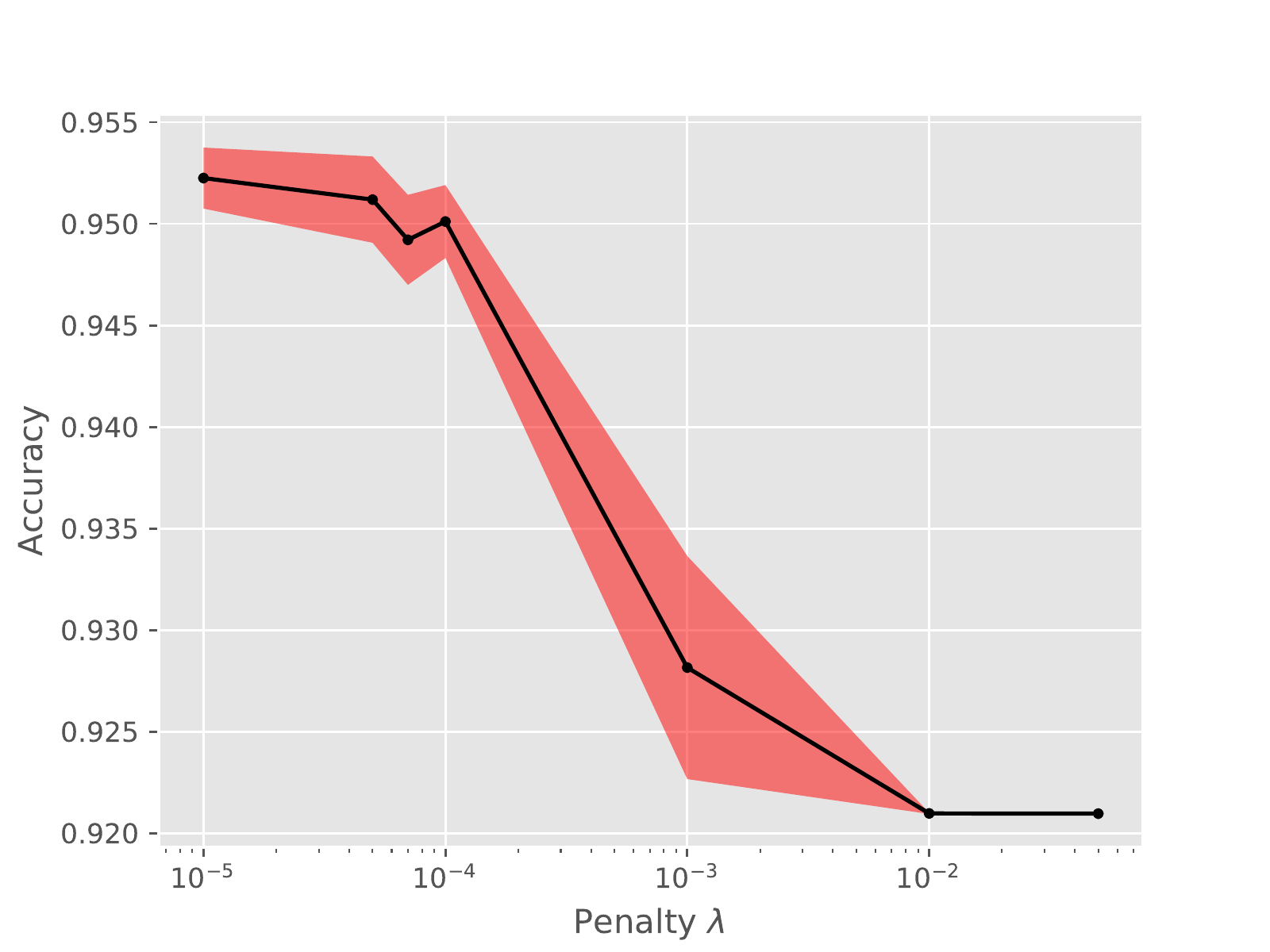}
\caption{Effect of varying $\lambda$ on CEM type system discovery}
\label{fig:lambda}
\end{figure}

\begin{figure*}[ht]
\centering
\begin{subfigure}{0.24\linewidth}
    \includegraphics[width=\linewidth]{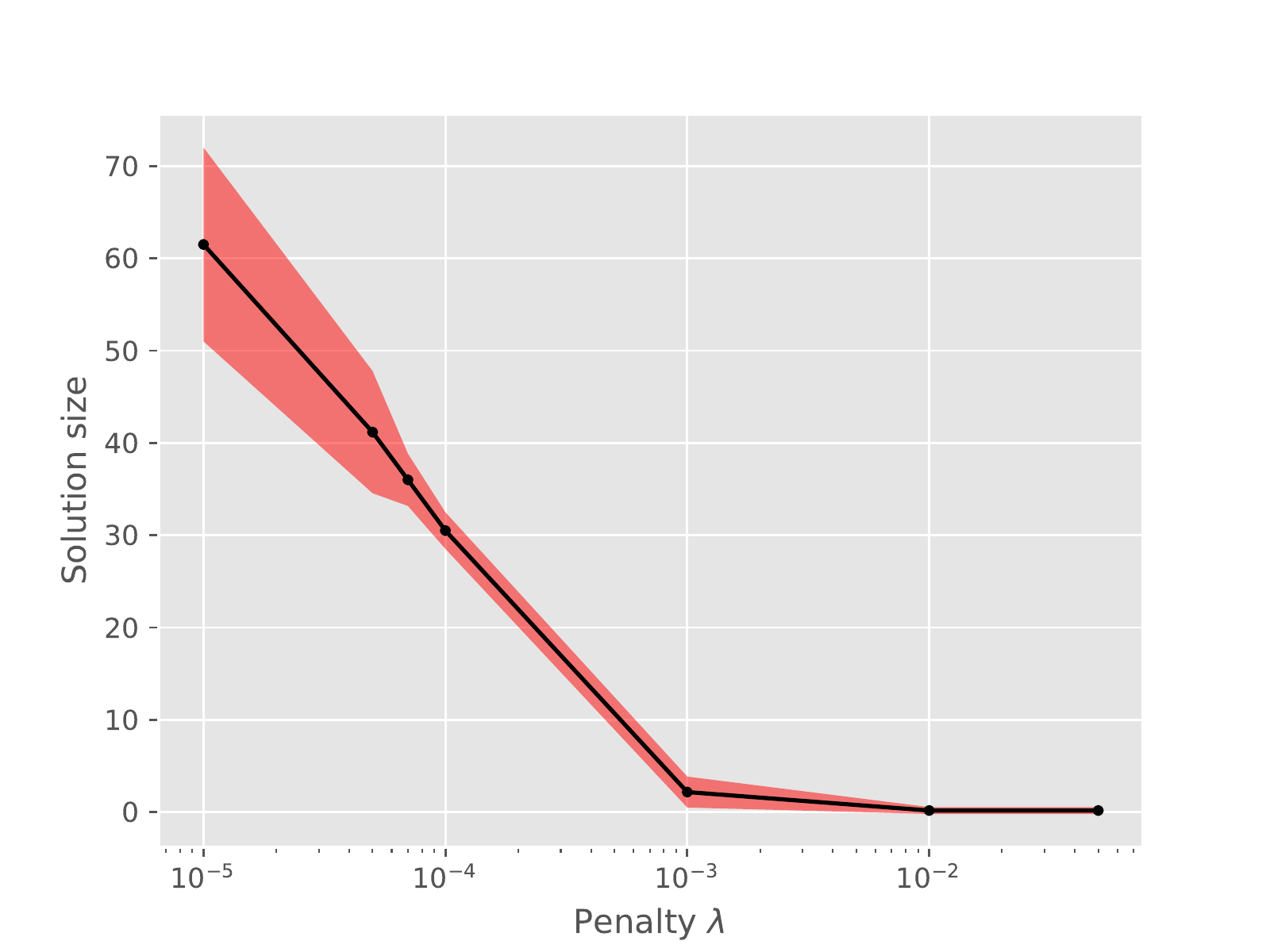}
    \caption{}
    \end{subfigure}
    \begin{subfigure}{0.24\linewidth}
    \includegraphics[width=\linewidth]{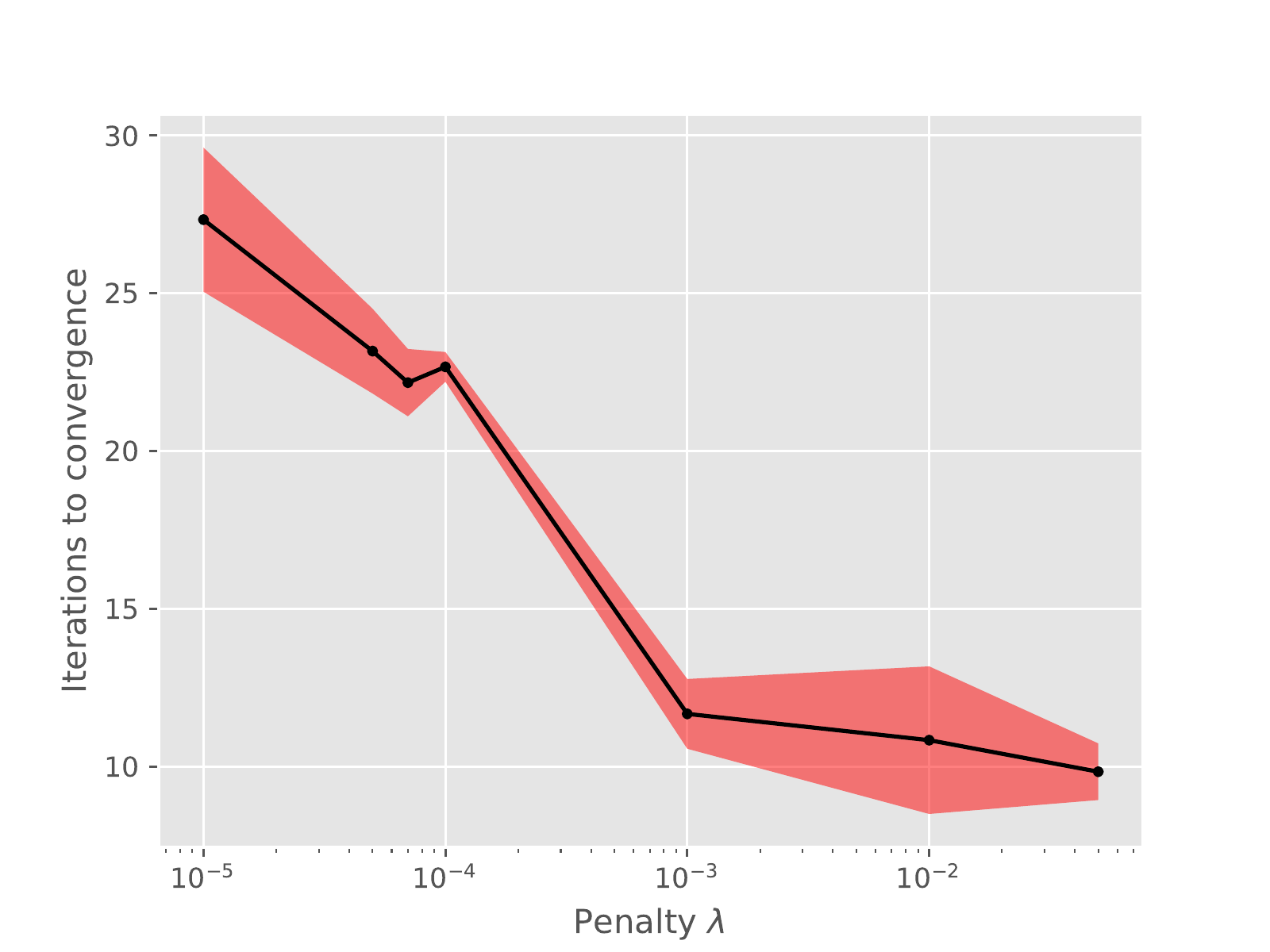}
    \caption{}
    \end{subfigure}
    \begin{subfigure}{0.24\linewidth}
    \includegraphics[width=\linewidth]{figures/lambda-score.pdf}
    \caption{}
    \end{subfigure}
    \begin{subfigure}{0.24\linewidth}
    \includegraphics[width=\linewidth]{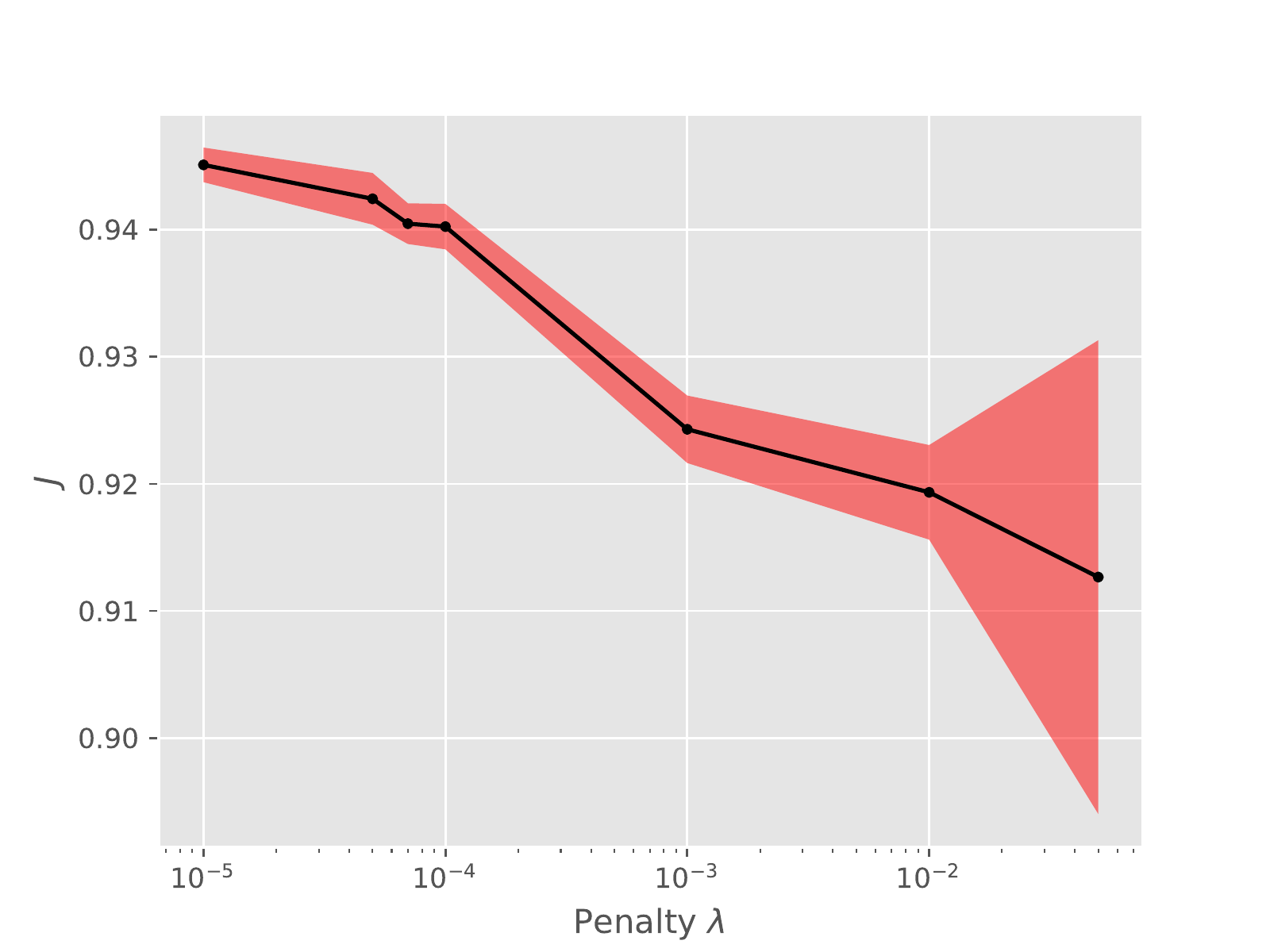}
    \caption{}
    \end{subfigure}
    \caption{Effect of varying $\lambda$ on CEM type system discovery: Solution size (a) and iterations to convergence (b) grow exponentially with penalty decrease, while accuracy plateaus (c) around $\lambda=10^{-4}$. Objective function increases as penalty decreases, since solution size is less penalized (d). Standard deviation is shown as the red region around the mean.}
    \label{fig:lambda}
\end{figure*}

We measure the effect of varying $\lambda$ on type system discovery when using CEM for our search. The effect averaged on 10 trials for a variety of $\lambda$ penalties is shown in Figure \ref{fig:lambda}. In particular we notice that there is a crossover point in the performance characteristics when selecting $\lambda$, where a looser penalty has diminishing returns in accuracy around $\lambda=10^{-4}$.

\section{Learnability Heuristic behavior}
\begin{figure*}[ht]
\centering
\begin{subfigure}{0.32\linewidth}
    \includegraphics[width=\linewidth]{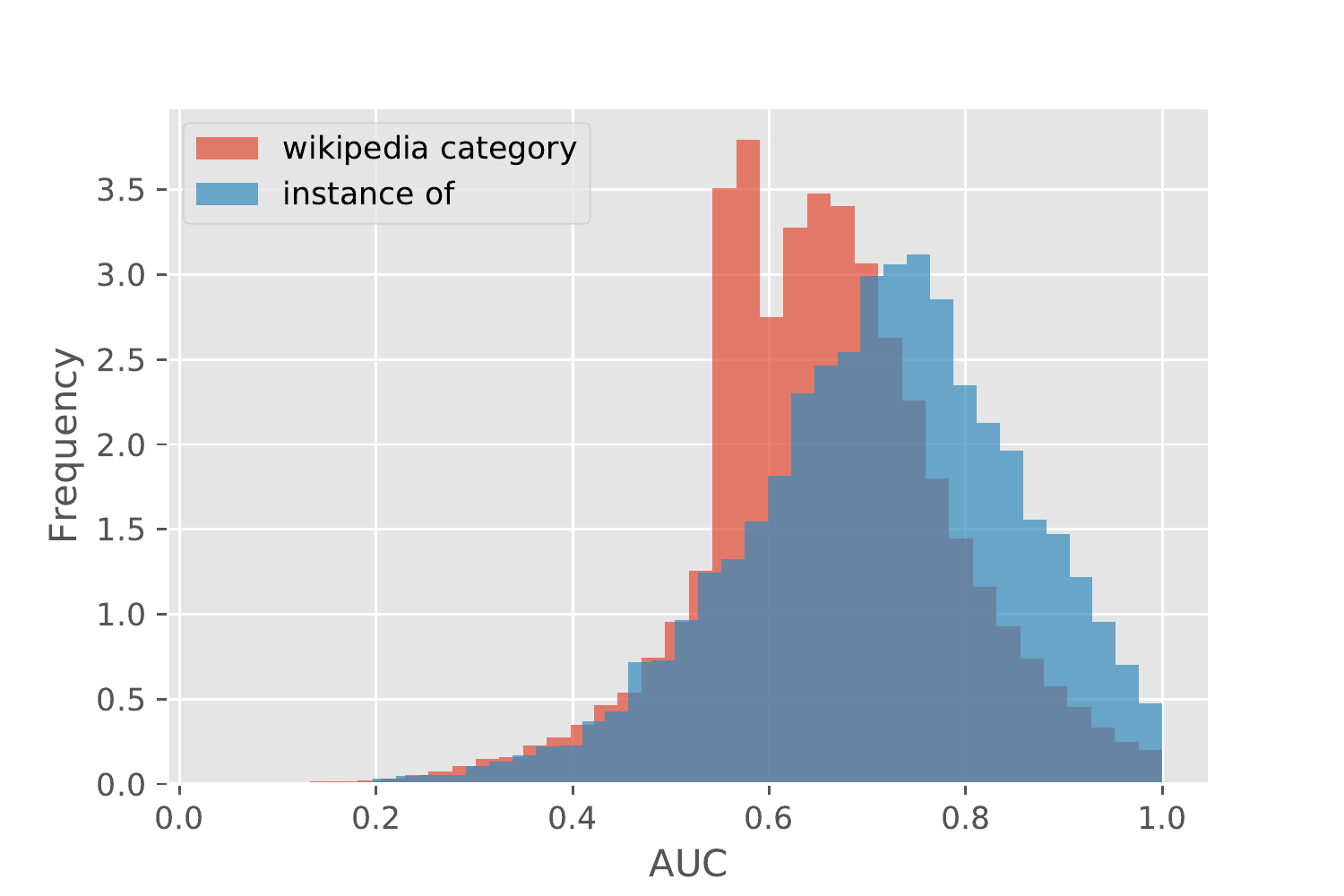}
    \caption{}
    \end{subfigure}
    \begin{subfigure}{0.32\linewidth}
    \includegraphics[width=\linewidth]{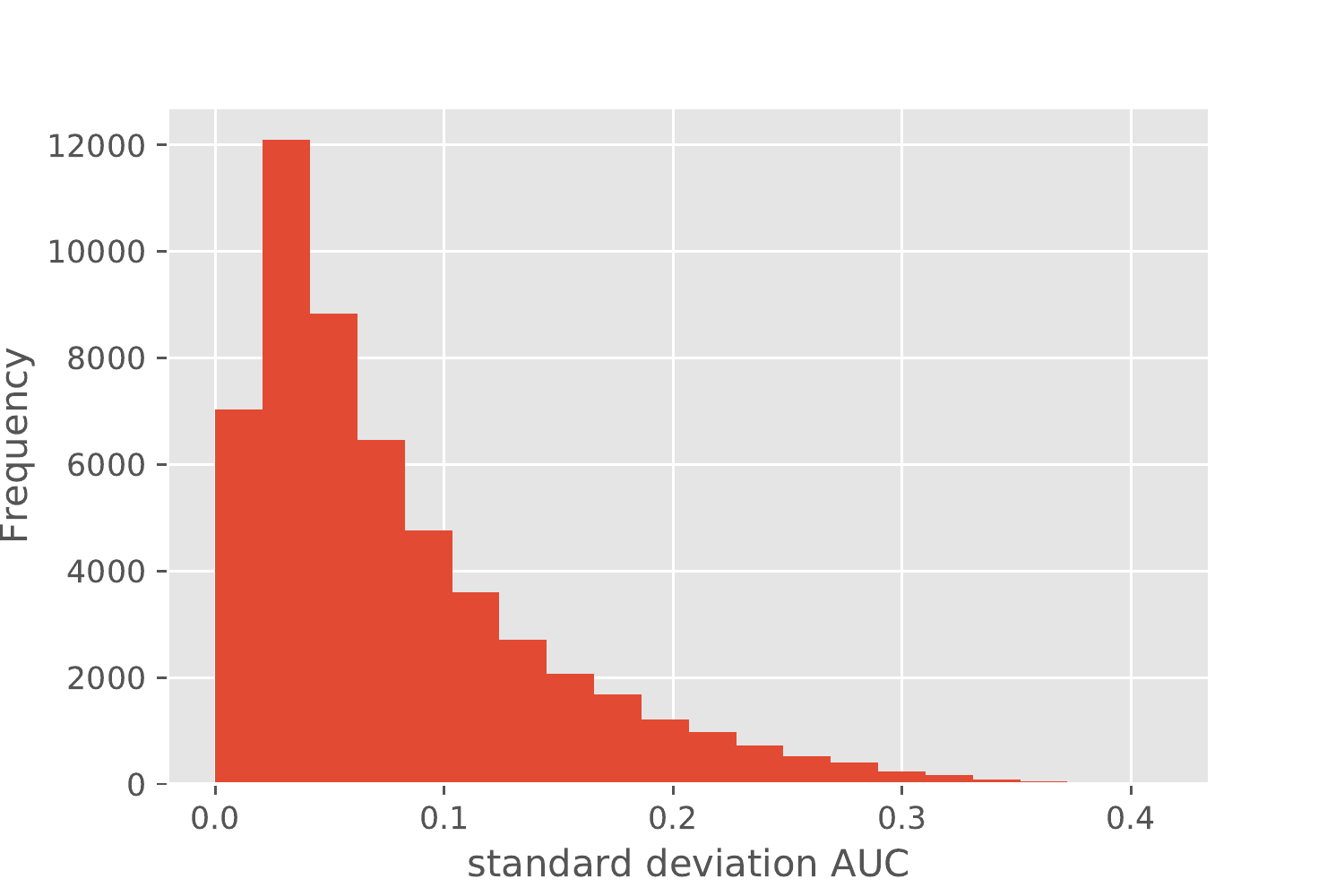}
    \caption{}
    \end{subfigure}
    \begin{subfigure}{0.32\linewidth}
    \includegraphics[width=\linewidth]{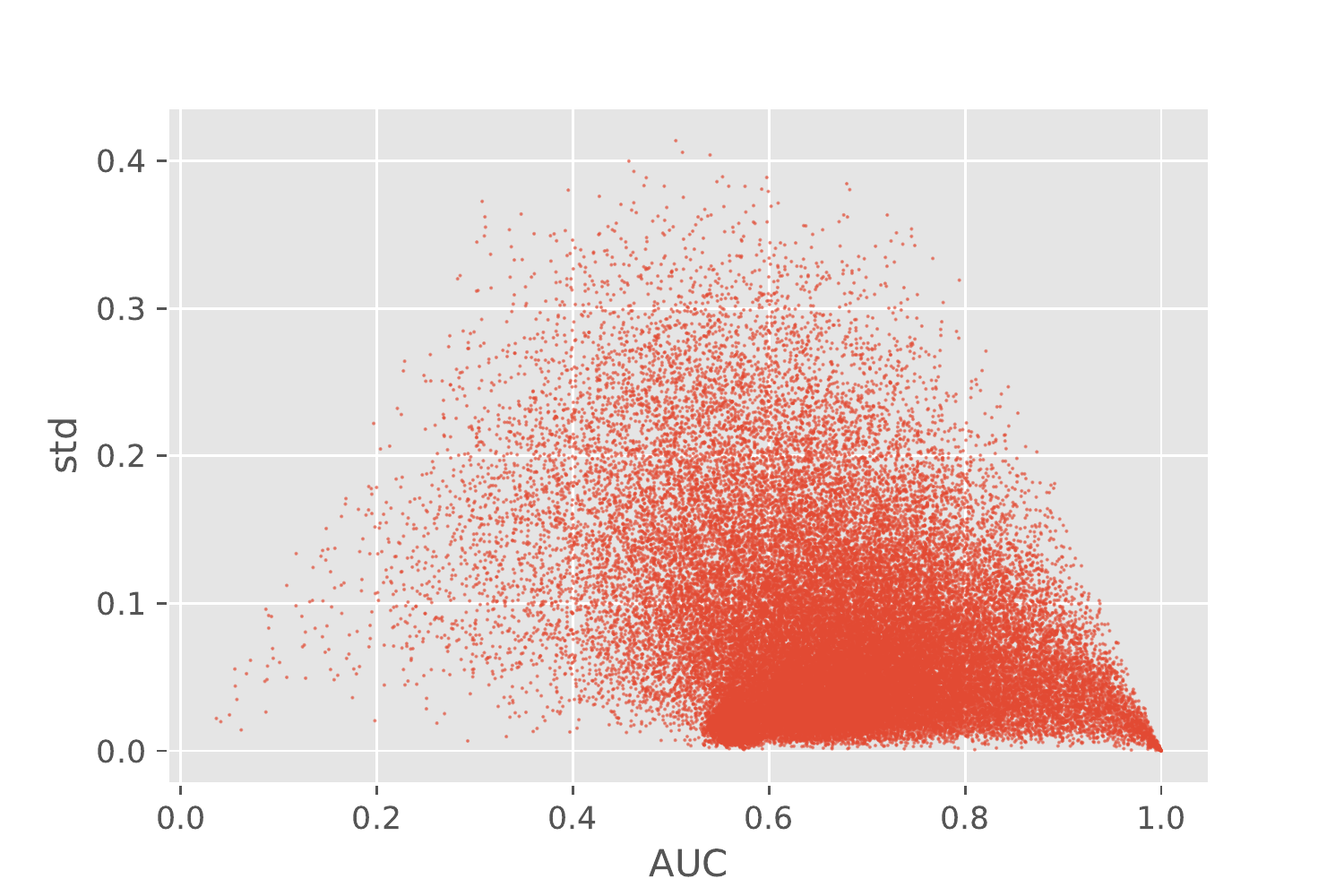}
    \caption{}
    \end{subfigure}
    \caption{Most instance of type-axes have higher AUC scores than wikipedia categories (a). The standard deviation for AUC scoring with text window classifiers is below 0.1 (b), AUC is not correlated with AUC's standard deviation.}
    \label{fig:learnability}
\end{figure*}

To better understand the behavior of the population of classifiers used to obtain AUC scores for the Learnability heuristic we investigate whether certain type axes are systematically easier or harder to predict, and summarize our results in Figure \ref{fig:learnability}. We find that type axes with a \texttt{instance of} edge have on average higher AUC scores than type axes relying on \texttt{wikipedia category}. Furthermore, we also wanted to ensure that our methodology for estimating learnability was not flawed or if variance in our measurement was correlated with AUC for a type axis. We find that there is no obvious relation between the standard deviation of the AUC scores for a type axis and the AUC score itself.

\section{Multilingual Part of Speech Tagging}

Finally the usage of multilingual data allows some amount of subjective experiments. For instance in Figure \ref{fig:multilingual} we show some samples from the model trained jointly on english and french correctly detecting the meaning of the word ``car" across three possible meanings.

\newcommand\replace[2]{\raisebox{-\baselineskip}{\shortstack{#1\\{\tiny #2}}}}
\begin{figure}[ht]
\fbox{\parbox{\textwidth}{
\begin{center}
\replace{This}{PRON} \replace{is}{VERB} \replace{a}{DET} \replace{\bf car}{\bf NOUN} \replace{,}{PCT} \replace{ceci}{PRON} \replace{n'}{PART} \replace{est}{VERB} \replace{pas}{ADV} \replace{une}{DET} \replace{voiture}{NOUN} \replace{\bf car}{\bf CONJ} \replace{c'}{PRON} \replace{est}{VERB} \replace{un}{DET} \replace{\bf car}{\bf NOUN} \replace{.}{PCT}
\end{center}
}}
\caption{Model trained jointly on monolingual POS corpora detecting the multiple meanings of ``car" (shown in bold) in a mixed English-French sentence.}
\label{fig:multilingual}
\end{figure}

\section{Human Type System}
\label{section:humandepth}
To assist humans with the design of the system, the rules are built interactively in a REPL, and execute over the 24 million entities in under 10 seconds, allowing for real time feedback in the form of statistics or error analysis over an evaluation corpus. On the evaluation corpus, disambiguation mistakes can be grouped according to the ground truth type, allowing a per type error analysis to easily detect areas where more granularity would help.
Shown below are the 5 different type axes designed by humans.
\begin{table*}[ht]
\caption{Human Type Axis: IsA}
\begin{center}
\begin{tabular}{|l|}\hline
Activity\\
Aircraft\\
Airport\\
Algorithm\\
Alphabet\\
Anatomical structure\\
Astronomical object\\
Audio visual work\\
Award\\
Award ceremony\\
Battle\\
Book magazine article\\
Brand\\
Bridge\\
Character\\
Chemical compound\\
Clothing\\
Color\\
Concept\\
Country\\
Crime\\
Currency\\
Data format\\
Date\\
Developmental biology period\\
Disease\\
Electromagnetic wave\\
Event\\
Facility\\
Family\\
Fictional character\\
Food\\
Gas\\
Gene\\
Genre\\
Geographical object\\
Geometric shape\\
Hazard\\
Human\\
Human female\\
Human male\\
International relations\\
\hline
\end{tabular}
\end{center}
\label{table:}
\end{table*}\begin{table*}[ht]
\caption{Human Type Axis: IsA (continued)}
\begin{center}
\begin{tabular}{|l|}\hline
Kinship\\
Lake\\
Language\\
Law\\
Legal action\\
Legal case\\
Legislative term\\
Mathematical object\\
Mind\\
Molecule\\
Monument\\
Mountain\\
Musical work\\
Name\\
Natural phenomenon\\
Number\\
Organization\\
Other art work\\
People\\
Person role\\
Physical object\\
Physical quantity\\
Plant\\
Populated place\\
Position\\
Postal code\\
Radio program\\
Railroad\\
Record chart\\
Region\\
Religion\\
Research\\
River\\
Road vehicle\\
Sea\\
Sexual orientation\\
Software\\
Song\\
Speech\\
Sport\\
Sport event\\
Sports terminology\\
Strategy\\
Taxon\\
Taxonomic rank\\
Title\\
Train station\\
Union\\
Unit of mass\\
\hline
\end{tabular}
\end{center}
\label{table:}
\end{table*}\begin{table*}[ht]
\caption{Human Type Axis: IsA (continued)}
\begin{center}
\begin{tabular}{|l|}\hline
Value\\
Vehicle\\
Vehicle brand\\
Volcano\\
War\\
Watercraft\\
Weapon\\
Website\\
Other\\
\hline
\end{tabular}
\end{center}
\label{table:}
\end{table*}\begin{table*}[ht]
\caption{Human Type Axis: Topic}
\begin{center}
\begin{tabular}{|l|}\hline
Archaeology\\
Automotive industry\\
Aviation\\
Biology\\
Botany\\
Business other\\
Construction\\
Culture\\
Culture-comics\\
Culture-dance\\
Culture-movie\\
Culture-music\\
Culture-painting\\
Culture-photography\\
Culture-sculpture\\
Culture-theatre\\
Culture arts other\\
Culture ceramic art\\
Culture circus\\
Culture literature \\
Economics\\
Education\\
Electronics\\
Energy\\
Engineering\\
Environment\\
Family\\
Fashion\\
Finance\\
Food\\
Health-alternative-medicine\\
Health-science-audiology\\
Health-science-biotechnology\\
Healthcare\\
Health cell\\
Health childbrith\\
Health drug\\
Health gene\\
Health hospital\\
Health human gene\\
Health insurance\\
Health life insurance\\
Health medical\\
Health med activism\\
Health med doctors\\
Health med society\\
Health organisations\\
Health people in health\\
Health pharma\\
Health protein\\
\hline
\end{tabular}
\end{center}
\label{table:}
\end{table*}\begin{table*}[ht]
\caption{Human Type Axis: Topic (continued)}
\begin{center}
\begin{tabular}{|l|}\hline
Health protein wkp\\
Health science medicine\\
Heavy industry\\
Home\\
Hortculture and gardening\\
Labour\\
Law\\
Media\\
Military war crime\\
Nature\\
Nature-ecology\\
Philosophy\\
Politics\\
Populated places\\
Religion\\
Retail other\\
Science other\\
Science-anthropology\\
Science-astronomy\\
Science-biophysics\\
Science-chemistry\\
Science-computer science\\
Science-geography\\
Science-geology\\
Science-history\\
Science-mathematics\\
Science-physics\\
Science-psychology\\
Science-social science other\\
Science chronology\\
Science histology\\
Science meteorology\\
Sex industry\\
Smoking\\
Sport-air-sport\\
Sport-american football\\
Sport-athletics\\
Sport-australian football\\
Sport-baseball\\
Sport-basketball\\
Sport-climbing\\
Sport-combat sport\\
Sport-cricket\\
Sport-cue sport\\
Sport-cycling\\
Sport-darts\\
Sport-dog-sport\\
Sport-equestrian sport\\
Sport-field hockey\\
Sport-golf\\
\hline
\end{tabular}
\end{center}
\label{table:}
\end{table*}\begin{table*}[ht]

\caption{Human Type Axis: Topic (continued)}
\begin{center}
\begin{tabular}{|l|}\hline
Sport-handball\\
Sport-ice hockey\\
Sport-mind sport\\
Sport-motor sport\\
Sport-multisports\\
Sport-other\\
Sport-racquet sport\\
Sport-rugby\\
Sport-shooting\\
Sport-soccer\\
Sport-strength-sport\\
Sport-swimming\\
Sport-volleyball\\
Sport-winter sport\\
Sport water sport\\
Toiletry\\
Tourism\\
Transportation\\
Other\\
\hline
\end{tabular}
\end{center}
\label{table:}
\end{table*}\begin{table*}[ht]

\caption{Human Type Axis: Time}
\begin{center}
\begin{tabular}{|l|}\hline
Post-1950\\
Pre-1950\\
Other\\
\hline
\end{tabular}
\end{center}
\label{table}

\end{table*}
\begin{table*}[ht]
\caption{Human Type Axis: Location}
\begin{center}
\begin{tabular}{|l|}\hline
Africa\\
Antarctica\\
Asia\\
Europe\\
Middle East\\
North America\\
Oceania\\
Outer Space\\
Populated place unlocalized\\
South America\\
Other\\
\hline
\end{tabular}
\end{center}
\label{table:}
\end{table*}

\bibliography{bibliography}
\bibliographystyle{aaai}
\end{document}